\documentclass[10pt,twocolumn,letterpaper]{article}

\usepackage[pagenumbers]{cvpr} %

\definecolor{cvprblue}{rgb}{0.21,0.49,0.74}
\usepackage[pagebackref,breaklinks,colorlinks,allcolors=cvprblue]{hyperref}
\usepackage[nameinlink,capitalize,noabbrev]{cleveref}
\usepackage{marvosym}
\usepackage{multirow}
\usepackage{float}
\def\paperID{31410} %
\def\confName{CVPR}
\def\confYear{2026}

\title{Structural Graph Probing of Vision--Language Models}

\author{
Haoyu He$^{*\, \dagger}$ \quad
Yue Zhuo$^{*\,\ddagger}$ \quad
Yu Zheng\textsuperscript{\Letter}$^{\ddagger}$ \quad
Qi R. Wang\textsuperscript{\Letter}$^{\dagger}$\\
$^{\dagger}$Northeastern University \quad
$^{\ddagger}$Massachusetts Institute of Technology\\
{\tt\small \{he.haoyu1, q.wang\}@northeastern.edu} \quad
{\tt\small \{joyzhuo, yu\_zheng\}@mit.edu}
}

\begin{document}
\maketitle
\def\thefootnote{*}
\footnotetext{Equal contribution.}

\begin{abstract}
    Vision--language models (VLMs) achieve strong multimodal performance, yet how computation is organized across populations of neurons remains poorly understood. In this work, we study VLMs through the lens of neural topology, representing each layer as a within-layer correlation graph derived from neuron--neuron co-activations. This view allows us to ask whether population-level structure is behaviorally meaningful, how it changes across modalities and depth, and whether it identifies causally influential internal components under intervention. We show that correlation topology carries recoverable behavioral signal; moreover, cross-modal structure progressively consolidates with depth around a compact set of recurrent hub neurons, whose targeted perturbation substantially alters model output. Neural topology thus emerges as a meaningful intermediate scale for VLM interpretability: richer than local attribution, more tractable than full circuit recovery, and empirically tied to multimodal behavior. Code is publicly available at \url{https://github.com/he-h/vlm-graph-probing}.
\end{abstract}

\section{Introduction}

Vision--language models (VLMs) have advanced rapidly as general-purpose multimodal systems \cite{AntolVQA, DBLP:journals/corr/abs-2103-00020, alayrac2022flamingo, li2023blip, zhu2023minigpt4enhancingvisionlanguageunderstanding}, yet how their competence is organized internally remains incompletely understood \cite{stan2024lvlm, dang2024explainable}. A central unresolved problem is how multimodal computation is organized within the network: how visual evidence and linguistic context are coordinated across layers, how intermediate computation is distributed across populations of units, and which internal structures ultimately govern model behavior. Without an account at that level, interpretability remains largely descriptive, useful for identifying salient inputs or components but limited in its ability to explain how multimodal reasoning is internally structured \cite{huang2024surveyevaluationmultimodallarge, yu2024understanding}.

Existing analyses of VLMs have primarily emphasized local explanatory signals, including attention patterns, saliency maps, patch attribution, and component-level inspection \cite{abnar2020quantifying, selvaraju2017grad, belinkov2022probing, DBLP:journals/corr/abs-1905-09418, yu2024understanding, neo2024towards}. While these approaches are valuable for identifying influential inputs or localized mechanisms, they are less well suited to characterizing the large-scale organization through which multimodal behavior is realized. This limitation is especially consequential in transformer-based VLMs, where computation is distributed across large populations of interacting units rather than concentrated in a small number of isolated pathways \cite{DBLP:journals/corr/abs-1908-07490, alayrac2022flamingo, li2023blip}. More broadly, both neuroscience and mechanistic interpretability point to a common lesson: complex computation often becomes most intelligible at the level of structured populations, interaction patterns, and hub-like organization, rather than at the level of individual units in isolation \cite{SpornsConnectome, olah2020zoom, elhage2021mathematical, BassettSmallWorld, HoneyConnectivity, cunningham2023sparseautoencodershighlyinterpretable}. For VLMs, this suggests that population-level interaction structure is not merely an auxiliary diagnostic, but a meaningful level of analysis in its own right \cite{kornblith2019similarity, zhang2023emergent, zheng2025probing}.

In this work, we study VLMs through the lens of \textit{neural topology}: the layerwise structure of neuron correlation induced during multimodal inference. Given an image--question pair, we record hidden activations, construct within-layer neuron correlation graphs, and use these graphs as a central object of analysis \cite{zheng2025probing}. This topology-centered perspective makes it possible to examine multimodal transformers at the level of population organization: to assess how strongly model behavior is reflected in internal interaction structure, to trace how cross-modal organization evolves across depth, and to identify structurally important neurons whose perturbation alters model predictions. Across multiple VLM families, we find that graph-based probes recover substantial signal about model outputs and hallucination behavior; that modality-specific topology reveals systematic changes in cross-modal organization across layers; and that perturbing topology-defined hub neurons significantly changes model predictions.

Taken together, these results establish neural topology as a meaningful perspective on vision--language model interpretability. Rather than treating multimodal interpretability solely as the problem of identifying salient tokens or image regions, our findings suggest that the organization of neuron correlations itself carries consequential information about model behavior. Under this view, \textit{behavioral predictability, multimodal structure, and intervention} are best understood not as separate analyses, but as complementary perspectives on the same underlying computational organization. This, in turn, points toward a broader approach to VLM interpretability centered on computation as an organized population-level system rather than as a collection of local attribution effects.

\begin{figure}[t]
    \centering
    
    \includegraphics[width=.9\linewidth]{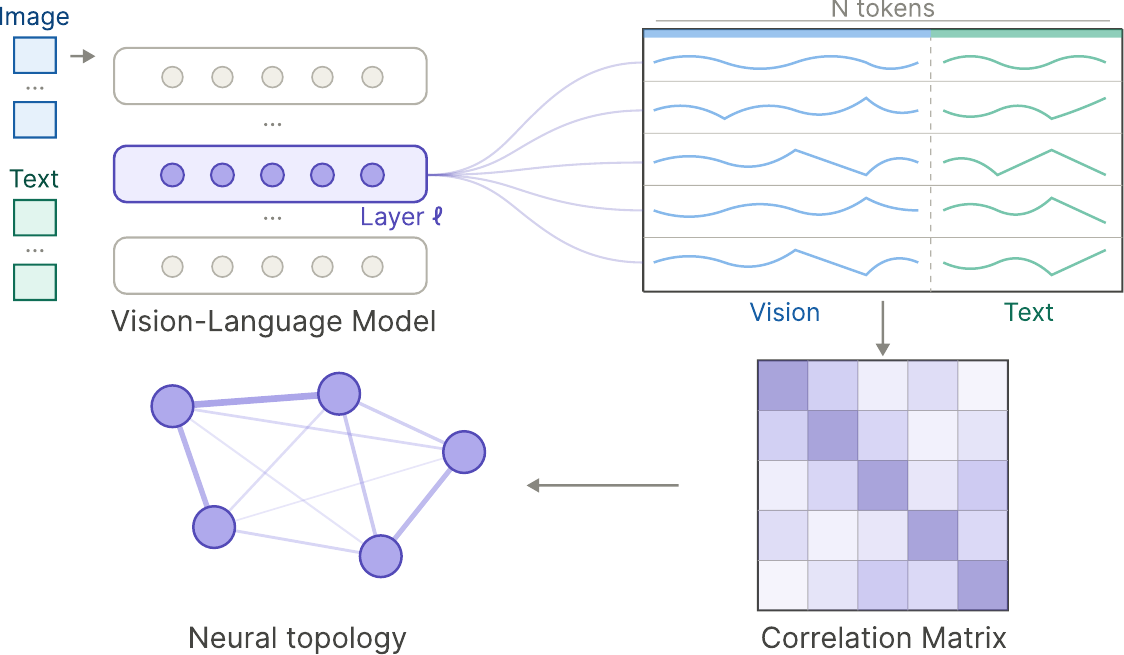}
    \caption{
        \textbf{Overview of neural topology construction.}
    }
    \label{fig:topology}
\end{figure}

\section{Neural Topology}

We analyze vision-language models at the level of \textit{population  structure}. Rather than treating multimodal reasoning as the output of isolated tokens, heads, or neurons, we view each transformer layer as a distributed computational state whose organization is expressed through the correlations among its units.
This choice is motivated by findings that functionally relevant organization is often expressed through distributed representational subspaces and heterogeneous network topology, including hub-dominated architectures, rather than through individual units in isolation \cite{doerig2025highlevel, piazza2025physical}.
In this work, we operationalize that perspective through \textit{neural correlation topology}, a layerwise description of neuron--neuron correlation structure during multimodal inference.

\subsection{Neuron Correlation Topology}
Given an image $I$ and text prompt $T$, a frozen VLM produces hidden activations at each transformer layer $\ell$. We denote the hidden representation at layer $\ell$ as $H^{(\ell)} \in \mathbb{R}^{d \times N}$,
where $N$ is the number of multimodal tokens and $d$ is the hidden dimension. Each row $H^{(\ell)}_{i,:}$ therefore represents the response of neuron $i$ across all tokens in the joint multimodal context. We use these hidden states only to infer correlation structure; the downstream analysis module never receives the activation values themselves.

We represent each layer as a weighted graph
\begin{equation}
G^{(\ell)} = (V, E, W^{(\ell)}),
\label{eq:graph}
\end{equation}
where each node in $V$ corresponds to a neuron, so that $|V| = d$, and $E = V \times V$ denotes the complete set of neuron pairs. The edge weight $W^{(\ell)}_{ij}$ measures the functional coupling between neurons $i$ and $j$, defined here as the Pearson correlation between their activation profiles across tokens:
\begin{equation}
W^{(\ell)}_{ij}
= \mathrm{corr}\!\left(H^{(\ell)}_{i,:}, H^{(\ell)}_{j,:}\right)
=
\frac{\left(H^{(\ell)}_{i,:}-\bar{H}^{(\ell)}_{i,:}\right)^\top
\left(H^{(\ell)}_{j,:}-\bar{H}^{(\ell)}_{j,:}\right)}
{\left\|H^{(\ell)}_{i,:}-\bar{H}^{(\ell)}_{i,:}\right\|
\left\|H^{(\ell)}_{j,:}-\bar{H}^{(\ell)}_{j,:}\right\|}.
\label{eq:corr}
\end{equation}
This construction yields a layerwise correlation graph in which edge weights reflect how similarly pairs of neurons respond within the same inference pass, shown in \Cref{fig:topology}. In other words, neural topology captures the organization of co-activation structure within each layer, a correlation-based view of within-layer population structure rather than a literal wiring diagram of the model.

\subsection{Vision, Text, and Multimodal Topology}
To examine how internal structure differs across modalities, we construct three related topologies at each layer. The multimodal graph $G^{(\ell)}$ is computed from the full hidden state obtained under joint image--text inference. To extract modality-specific structure, we use the same multimodal forward pass and partition the hidden states by token type, using positional indices to separate visual and textual tokens. Let $H^{(\ell)}_{\mathrm{vis}}$ and $H^{(\ell)}_{\mathrm{text}}$ denote the subsets of hidden activations associated with vision and text tokens, respectively. We then construct
\begin{equation}
G^{(\ell)}_{\mathrm{vis}} = \mathrm{GraphCorr}\!\left(H^{(\ell)}_{\mathrm{vis}}\right),
G^{(\ell)}_{\mathrm{text}} = \mathrm{GraphCorr}\!\left(H^{(\ell)}_{\mathrm{text}}\right),
\label{eq:modality_graphs}
\end{equation}
using the same correlation-based procedure as in the multimodal case. Because these graphs are derived from different token subsets within the same forward pass, differences among $G^{(\ell)}$, $G^{(\ell)}_{\mathrm{vis}}$, and $G^{(\ell)}_{\mathrm{text}}$ reflect how correlation structure specializes to visual tokens, textual tokens, and their joint multimodal context.

\begin{table*}[t]
\centering
\small
\caption{
\textbf{Graph probing across multimodal benchmarks.}
Accuracy and F1 of linear vs. graph-based probes trained on layer-wise neuron-correlation topology for TDIUC, CLEVR, MMMU, MMMU-Pro, BLINK, and EMMA. Best result in each model–dataset pair is in \textbf{bold}. Graph-based probes outperform linear baselines in most settings, indicating that population-level connectivity carries task-relevant information. See \Cref{tab:vlm_cls_pr} for Precision/Recall.
}
\label{tab:vlm_cls}
\begin{tabular}{lcccccccccccc}
\toprule
& \multicolumn{4}{c}{\textbf{InternVL3-1B}} 
& \multicolumn{4}{c}{\textbf{Qwen2.5-VL-3B}} 
& \multicolumn{4}{c}{\textbf{LLaVA-1.5-7B}} \\
\cmidrule(lr){2-5} \cmidrule(lr){6-9} \cmidrule(lr){10-13}
\textbf{Dataset}
& \multicolumn{2}{c}{\textbf{Linear}} & \multicolumn{2}{c}{\textbf{GCN}}
& \multicolumn{2}{c}{\textbf{Linear}} & \multicolumn{2}{c}{\textbf{GCN}}
& \multicolumn{2}{c}{\textbf{Linear}} & \multicolumn{2}{c}{\textbf{GCN}} \\
\cmidrule(lr){2-3} \cmidrule(lr){4-5}
\cmidrule(lr){6-7} \cmidrule(lr){8-9}
\cmidrule(lr){10-11} \cmidrule(lr){12-13}
& Acc & F1 & Acc & F1 & Acc & F1 & Acc & F1 & Acc & F1 & Acc & F1 \\
\midrule
TDIUC
& 0.884 & 0.846 & \textbf{0.965} & \textbf{0.959}
& 0.943 & 0.930 & \textbf{0.976} & \textbf{0.970}
& \textbf{0.971} & \textbf{0.966} & 0.954 & 0.947 \\
CLEVR
& 0.980 & 0.980 & \textbf{0.993} & \textbf{0.993}
& 0.920 & 0.919 & \textbf{0.963} & \textbf{0.963}
& 0.602 & 0.563 & \textbf{0.679} & \textbf{0.676} \\
MMMU
& 0.293 & \textbf{0.253} & \textbf{0.321} & \textbf{0.253}
& 0.293 & 0.211 & \textbf{0.336} & \textbf{0.335}
& \textbf{0.314} & \textbf{0.300} & 0.279 & 0.162 \\
MMMU-Pro
& 0.309 & 0.226 & \textbf{0.359} & \textbf{0.260}
& 0.286 & 0.164 & \textbf{0.318} & \textbf{0.290}
& 0.291 & 0.171 & \textbf{0.314} & \textbf{0.305} \\
BLINK
& 0.549 & 0.530 & \textbf{0.592} & \textbf{0.589}
& 0.544 & 0.543 & \textbf{0.565} & \textbf{0.564}
& \textbf{0.647} & \textbf{0.646} & 0.592 & 0.591 \\
EMMA
& 0.343 & 0.195 & \textbf{0.380} & \textbf{0.268}
& 0.329 & 0.146 & \textbf{0.343} & \textbf{0.288}
& 0.307 & \textbf{0.228} & \textbf{0.329} & 0.218 \\
\bottomrule
\end{tabular}
\end{table*}

\subsection{Neural Topology Representation}
To analyze these graphs without directly exposing hidden activation magnitudes or token semantics, we represent each neuron using a learnable one-hot node identity embedding. Each neuron is assigned a unique basis vector, which is projected by a trainable embedding layer into a low-dimensional feature space. The resulting node features preserve neuron identity while keeping the analysis centered on correlation structure rather than raw hidden-state content. A graph convolutional network (GCN) operates over the correlation graph to produce topology-dependent node representations:
\begin{equation}
Z^{(\ell)} = \mathrm{GCN}\!\left(W^{(\ell)}, X\right)
= \sigma\!\left(D^{-\frac{1}{2}} W^{(\ell)} D^{-\frac{1}{2}} X W_g\right),
\label{eq:gcn}
\end{equation}
where $X$ denotes the node embedding matrix, $D$ is the degree matrix of $W^{(\ell)}$, $W_g$ is a learnable weight matrix, and $\sigma(\cdot)$ is a nonlinear activation function. Since the GCN receives only graph structure and node identities, the resulting representation reflects how neurons are organized within the layer rather than what individual neurons encode.

To obtain a fixed-dimensional structural signature for each layer, we aggregate the node representations with complementary global pooling operators:
\begin{equation}
h^{(\ell)} =
\mathrm{Concat}\!\left(
\mathrm{Mean}\!\left(Z^{(\ell)}\right),
\mathrm{Max}\!\left(Z^{(\ell)}\right)
\right).
\label{eq:pool}
\end{equation}
The mean-pooled term captures the overall correlation tendency of the layer, while the max-pooled term preserves salient high-response structure. Each transformer layer therefore yields a graph-level signature $h^{(\ell)}$ that can be used to study behavioral predictability, multimodal structure, and intervention targets in a common representation space.

\section{Predictability}

A central question is whether neural topology encodes behaviorally meaningful information, rather than merely reflecting incidental patterns of co-activation. We address this question by asking whether layerwise correlation graphs support reliable prediction of model behavior across grounded reasoning, semantic recognition, and hallucination classification tasks. This section serves as the first empirical test of our framework: rather than claiming mechanism from probe performance alone, we use predictability to establish that neural topology is a structured and behaviorally informative representation, thereby motivating the more detailed structural and causal analyses that follow.

\paragraph{Experimental Setup.} We evaluate three representative VLMs: InternVL3-1B~\cite{zhu2025internvl3}, Qwen2.5-VL-3B~\cite{bai2025qwen2}, and LLaVA-1.5-7B~\cite{liu2023visual}. In the main text, we focus on CLEVR~\cite{johnson2017clevr}, TDIUC~\cite{kafle2017analysis}, and MHaluBench~\cite{chen2024unified}, which respectively test numerical grounding, semantic recognition, and multimodal hallucination. Broader results on MMMU~\cite{yue2024mmmu}, MMMU-Pro~\cite{yue2025mmmu}, BLINK~\cite{fu2024blink}, and EMMA~\cite{hao2025can} are reported in \Cref{tab:vlm_cls}. For each dataset, we split examples into 80\% training and 20\% test sets, and train both a linear probe and a GCN probe on each layer's graph representation. Full dataset descriptions, model details, and training settings are provided in \Cref{app:setup}.

\subsection{Behavioral Predictability}
\label{sec:probing}

We use CLEVR object counting to probe quantitative reasoning, and color recognition on both CLEVR and TDIUC alongside TDIUC sports classification to probe semantic understanding. These tasks allow us to test whether graph-derived representations of internal activity encode information that is predictive of model behavior.

\paragraph{Probing Performance.} \Cref{tab:vlm_cls} reports out-of-sample probing accuracy for linear and GCN probes across CLEVR and TDIUC, together with broader results on MMMU, MMMU-Pro, BLINK, and EMMA. Overall, graph-based probes outperform linear baselines on most model--dataset pairs, with the clearest gains on CLEVR counting and TDIUC. The strongest improvements appear on CLEVR, where the GCN probe improves over the linear baseline by $7.7\%$ on LLaVA, $4.3\%$ on Qwen2.5-VL, and $1.3\%$ on InternVL3. Broader multimodal benchmarks exhibit a more mixed pattern, suggesting that topology is especially informative on grounded tasks with tighter alignment between internal multimodal organization and target outputs.

These results indicate that modeling relational structure within the graph yields additional predictive value beyond a linear readout of the same graph-derived representation. Notably, all results are obtained on sparse correlation graphs with density at most $0.2$, showing that high predictive performance does not require dense connectivity.

Beyond classification, \Cref{tab:vlm_reg} evaluates CLEVR object counting as a regression problem. Across all three models, graph-based probes reduce MSE and improve both $R^2$ and Pearson correlation, indicating that the probe captures genuine functional dependence between topology and target counts rather than overfitting to dataset idiosyncrasies. This shows that the advantage of topology extends beyond discrete label prediction to finer-grained numerical estimation. Taken together, these results establish that neural topology contains recoverable signal about both semantic and quantitative task behavior.

\begin{table}[t]
\centering
\small
\caption{
\textbf{Regression on object counting from neural topology.}
Linear and graph-based probes on CLEVR counting.  Graph-based probes improve regression performance across all three VLMs.
}
\label{tab:vlm_reg}
\begin{tabular}{l l ccc}
\toprule
\textbf{Model} & \textbf{Probe} & \textbf{MSE $\downarrow$} & \textbf{$R^2$ $\uparrow$} & \textbf{Pearson $\uparrow$} \\
\midrule
\multirow{2}{*}{InternVL3-1B}
  & Linear & 0.020 & 0.996 & 0.998 \\
  & GCN    & \textbf{0.007} & \textbf{0.999} & \textbf{0.999} \\
\midrule
\multirow{2}{*}{Qwen2.5-VL-3B}
  & Linear & 0.081 & 0.985 & 0.992 \\
  & GCN    & \textbf{0.038} & \textbf{0.993} & \textbf{0.996} \\
\midrule
\multirow{2}{*}{LLaVA-1.5-7B}
  & Linear & 0.605 & 0.884 & 0.949 \\
  & GCN    & \textbf{0.379} & \textbf{0.928} & \textbf{0.963} \\
\bottomrule
\end{tabular}
\end{table}
\begin{figure}[t]
    \centering
    \includegraphics[width=\linewidth]{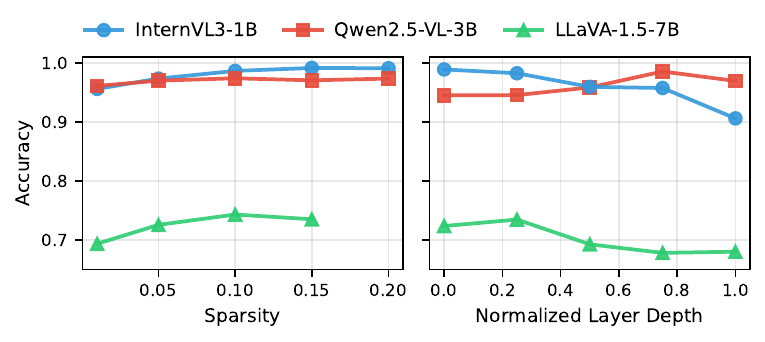}
    \caption{
        \textbf{Sparsity robustness and depth dependence of graph probing.}
Left: probing accuracy as a function of graph sparsity (top 1\%–20\% of neuron correlations retained). Right: probing accuracy across normalized layer depth. Accuracy is stable across sparsity levels, while depth-wise peak predictiveness differs by architecture.
    }
    \label{fig:sparsity_layer}
\end{figure}

\paragraph{Sparsity and Graph Construction.} A practical challenge in graph probing for VLMs is scale: each layer contains thousands of neurons, producing correlation graphs with tens of millions of possible edges. Fully connected graphs are computationally prohibitive and can obscure the strongest structural relations. To address this, a sparse construction strategy retains only the top-$k$ fraction of edges with the largest pairwise correlations, yielding a weighted graph that remains tractable while preserving dominant dependencies.

To evaluate the effect of this design choice, we sweep sparsity levels from $0.01$ to $0.20$. As shown in \Cref{fig:sparsity_layer}(a), probe accuracy remains largely stable across this range, with only marginal gains as additional weaker edges are included. This suggests that, under our probing setup, the most predictive structural signal is already concentrated in the strongest correlations. From a practical perspective, this justifies sparse graph construction as an efficient approximation; from an analytical perspective, it indicates that task-relevant topology can be recovered without modeling the full dense correlation graph.

\paragraph{Layerwise Predictability.} To examine how predictive topology varies with depth, probe accuracy is evaluated across layers of each VLM in \Cref{fig:sparsity_layer}(b). Qwen2.5-VL-3B exhibits a clear mid-to-late peak, reaching its highest accuracy around layer 27 before slightly declining near the final layer. In contrast, LLaVA-1.5-7B and InternVL3-1B show flatter or gradually declining trends across depth.

These results suggest that the depth at which topology is most behaviorally informative differs across architectures. Rather than drawing a strong functional conclusion from probe accuracy alone, these layerwise differences serve as motivation for the structural analysis in the next section, where we examine how multimodal correlation patterns evolve across depth.

\subsection{Hallucination Detection}
\label{sec:hallucination}

We test whether neural topology can distinguish hallucinating from non-hallucinating responses on MHaluBench. A binary classifier is trained on graph representations extracted from each model to predict hallucination status. As text-only controls, we construct two simple baselines using \texttt{word2vec}~\cite{mikolov2013efficientestimationwordrepresentations, rehurek2010gensim}: the mean embedding of each question–answer prompt. Separate linear probes are trained on these features to estimate how much hallucination is predictable from shallow textual statistics alone.

As shown in \Cref{tab:hallucination}, graph-based probes consistently outperform these text-only baselines across all three models, indicating that hallucination status is associated with structural information in neuron–neuron correlation graphs beyond simple lexical cues. This result is best viewed as evidence of informativeness rather than as a competitive hallucination-detection system: the main takeaway is that topological representations capture signals related to whether a response is grounded or hallucinatory.

\begin{table}[t]
\centering
\small
\caption{
\textbf{Hallucination detection from neural topology on MHaluBench.}
Accuracy of a graph-based probe versus two text-only baselines (mean word2vec embedding and token length) for binary hallucination classification. Graph-based probes outperform both baselines across all models.
}
\label{tab:hallucination}
\resizebox{\linewidth}{!}{%
\begin{tabular}{lccc}
\toprule
\textbf{Method} & \textbf{InternVL3-1B} & \textbf{Qwen2.5-VL-3B} & \textbf{LLaVA-1.5-7B} \\
\midrule
Emb.~Avg. & 0.664 & 0.654 & 0.649 \\
Length    & 0.500 & 0.633 & 0.642 \\
\midrule
GCN       & \textbf{0.789} & \textbf{0.910} & \textbf{0.908} \\
\bottomrule
\end{tabular}%
}
\end{table}

\begin{figure*}[t]
    \centering
    \includegraphics[width=.85\linewidth]{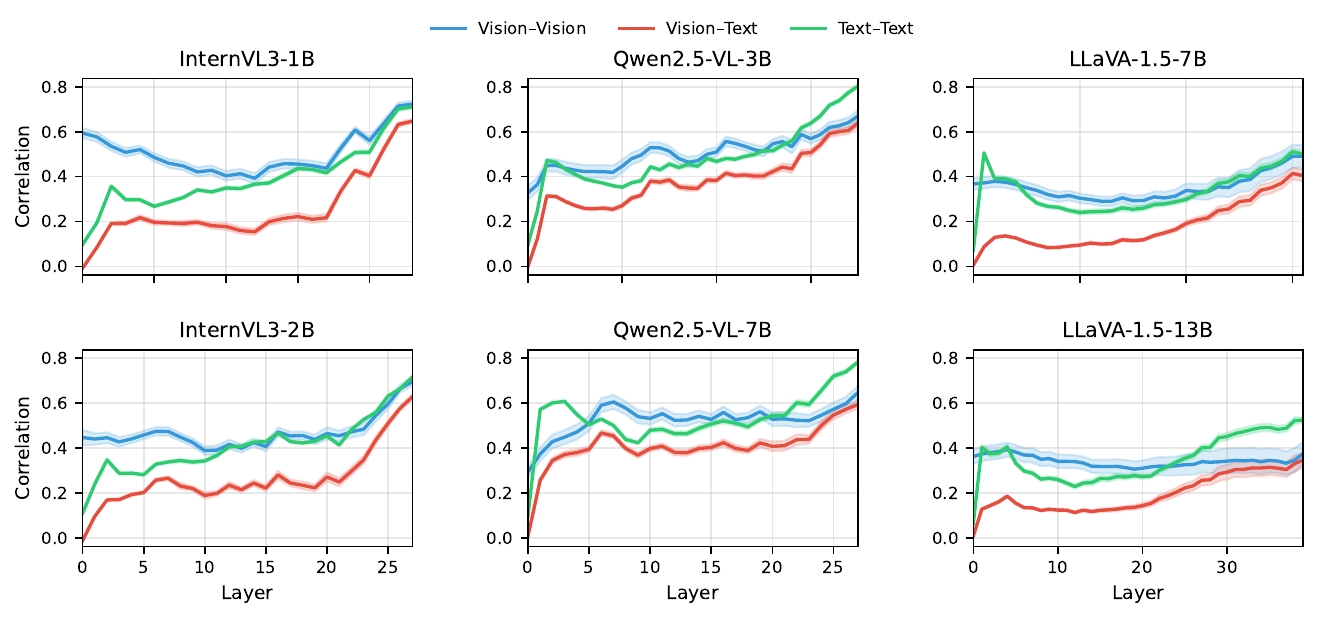}
    \caption{
        \textbf{Token-level cross-modal correlation dynamics across depth.}
        Layer-wise token–token correlations for Vision–Vision, Vision–Text, and Text–Text pairs on TDIUC (mean $\pm$ std) across multiple VLM families and scales. Vision–Text correlations increase with depth, consistent with progressively stronger multimodal integration in later layers.
    }
    \label{fig:modality_correlation}
\end{figure*}

\section{Multimodal Structure and Alignment}

Beyond predictability, a key question is what kind of multimodal organization neural topology reflects. We examine the hidden states at three complementary levels: token-level correlation dynamics, neuron-level persistence of structural roles, and graph-level alignment across modality conditions. Together, these analyses characterize how visual and linguistic information become coupled, stabilized, and organized within the internal topology of VLMs.

\subsection{Cross-Modal Correlation Dynamics}

We begin with a token-level view of multimodal interaction. Let $H^{(\ell)}\!\in\!\mathbb{R}^{d\times N}$ denote the hidden activations at layer~$\ell$, where $d$ is the hidden dimension and $N$ is the number of multimodal tokens. To quantify token--token dependencies, we compute the Pearson correlation of the transposed activations $C_{\text{tok}}^{(\ell)} = \mathrm{corr}\big(H^{(\ell)\top}\big)$.
For vision and text token sets $\mathcal{V}$ and $\mathcal{T}$, we define intra- and inter-modality coupling as
\begin{equation}
\begin{aligned}
\mu_{\mathrm{VV}}^{(\ell)} &=
\frac{1}{|\mathcal{V}|(|\mathcal{V}|-1)}
\sum_{\substack{i,j\in\mathcal{V}}} C_{\text{tok}}^{(\ell)}[i,j], \\[3pt]
\mu_{\mathrm{TT}}^{(\ell)} &=
\frac{1}{|\mathcal{T}|(|\mathcal{T}|-1)}
\sum_{\substack{i,j\in\mathcal{T}}} C_{\text{tok}}^{(\ell)}[i,j], \\[3pt]
\mu_{\mathrm{VT}}^{(\ell)} &=
\frac{1}{|\mathcal{V}|\,|\mathcal{T}|}
\sum_{i\in\mathcal{V}}\sum_{j\in\mathcal{T}} C_{\text{tok}}^{(\ell)}[i,j].
\end{aligned}
\label{eq:modality_coupling}
\end{equation}

As shown in \Cref{fig:modality_correlation}, both vision--text and text--text correlations increase with depth, whereas vision--vision correlations remain comparatively flat. This pattern is consistent with progressively stronger multimodal integration in later layers, although the statistic itself is descriptive rather than mechanistic. In decoder-style VLMs, this asymmetry may reflect the role of visual tokens as conditioning inputs that increasingly shape the language-side representation as depth increases.

\begin{figure*}[t!]
    \centering
    \includegraphics[width=.85\linewidth]{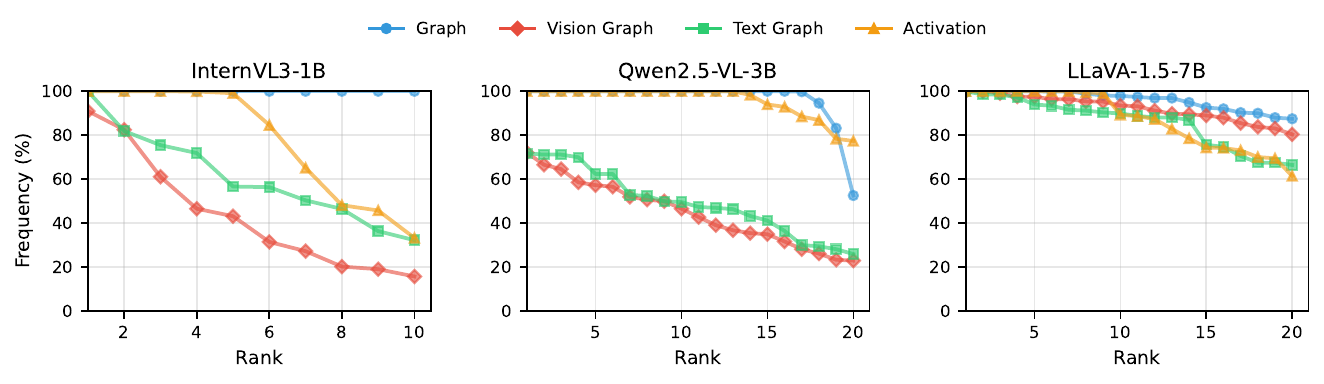}
    \caption{\textbf{Cross-sample stability of hub definitions.}
    Recurrence of top 1\% hubs across samples on TDIUC for graph-wide, modality-specific, and activation-based hub definitions. Graph-derived hubs are the most stable, indicating that structurally central neurons occupy more persistent roles than alternative hub candidates.
            }
    \label{fig:hub_neurons}
\end{figure*}
\begin{figure*}[t!]
    \centering
    \includegraphics[width=.85\linewidth]{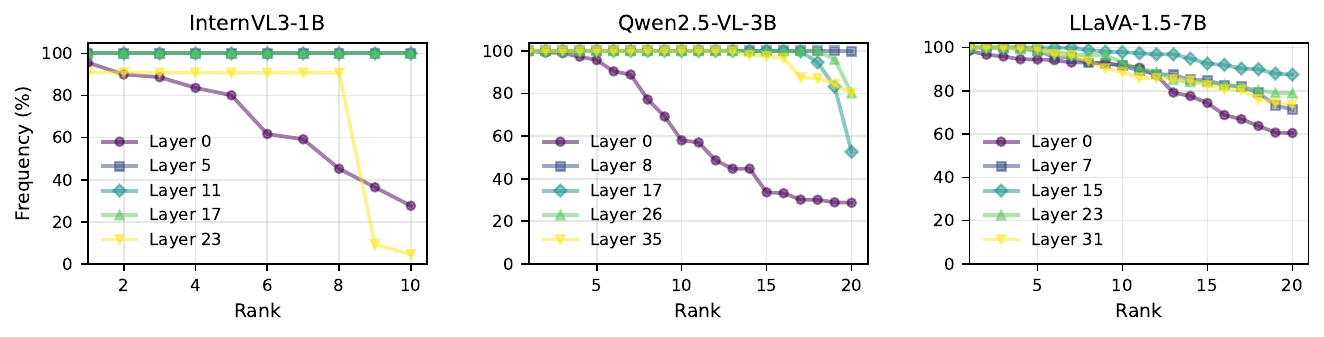}
    \caption{\textbf{Layer-wise stability of graph hubs.}
    Recurrence of top 1\% graph-derived hubs across samples at different depths on TDIUC. Intermediate layers show the strongest hub stability, suggesting the most persistent population-level organization emerges in mid-depth representations.
    }
    \label{fig:hub_neurons_layers}
\end{figure*}
\subsection{Structural Hub Stability}

A natural follow-up is whether topology identifies structural roles that persist across inputs. For each correlation graph $W^{(\ell)}$, we define the degree of neuron~$i$ as 
\begin{equation}
d_i^{(\ell)} = \sum_j |W_{ij}^{(\ell)}|.
\end{equation}
Neurons within the top-$k\%$ of $d_i^{(\ell)}$ are treated as \emph{hub neurons}. For a set of samples $\mathcal{S}$ with corresponding hub sets $\mathcal{H}_s^{(\ell)}$, we measure cross-sample recurrence as
\begin{equation}
\pi_i^{(\ell)} = \tfrac{1}{|\mathcal{S}|} \sum_{s\in\mathcal{S}} \mathbf{1}[i\in\mathcal{H}_s^{(\ell)}].
\end{equation}

The distributions in \Cref{fig:hub_neurons} show that graph-defined hubs exhibit substantially greater recurrence across samples than activation-based and modality-specific alternatives. This comparison disentangles three possible sources of hub persistence: full multimodal topology, unimodal subnetwork structure, and activation magnitude. Graph-defined hubs are the most stable, indicating that topology captures a more persistent notion of structural centrality than either alternative.

Stability further peaks in the middle layers (\Cref{fig:hub_neurons_layers}), broadly aligning with the region where cross-modal coupling becomes strongest. Rather than taking this as direct evidence of mediation, we interpret it as a compact set of structurally central neurons appears to recur reliably across diverse inputs, suggesting that multimodal processing is organized around persistent topological loci rather than being uniformly distributed across the network.

\subsection{Cross-Modal Graph Alignment}

The final analysis tests whether modality-specific graphs occupy a shared structural space. For each layer $\ell$, we extract graph-level embeddings $h_{\Omega}^{(\ell)}$ and $h_{\Gamma}^{(\ell)}$ from the GCN representations under different modality conditions. Rather than performing explicit node matching, we align these graph embeddings contrastively: positive pairs are drawn from the same sample and layer, while negative pairs are drawn from different samples or layers. We train the alignment model using a symmetric InfoNCE objective with cosine similarity and temperature $\tau$:
\begin{equation}
\mathcal{L} = -\tfrac{1}{2|\mathcal{B}|} \sum_{i\in\mathcal{B}}
\Big[\log\frac{e^{s(z_{\Omega,i},z_{\Gamma,i})/\tau}}
{\sum_j e^{s(z_{\Omega,i},z_{\Gamma,j})/\tau}} +
\log\frac{e^{s(z_{\Gamma,i},z_{\Omega,i})/\tau}}
{\sum_j e^{s(z_{\Gamma,i},z_{\Omega,j})/\tau}}\Big].
\label{eq:contrastive_alignment}
\end{equation}
We evaluate alignment using Graph AUC (GAUC)~\cite{li2019graphmatchingnetworkslearning}, which measures how reliably matched graph representations are ranked above mismatched ones.

As shown in \Cref{tab:matching}, multimodal--multimodal matching provides the highest alignment score (0.9598 GAUC), serving as a reference point for near self-alignment within the learned structural space. Matching LLaVA's text and image pathways yields a lower score (0.8188), indicating that unimodal graphs derived from the same multimodal model remain only partially aligned. Alignment drops further when comparing LLaVA text graphs with those from the unimodal LLaMA backbone~\cite{touvron2023llama2openfoundation} (0.6803), suggesting that multimodal finetuning alters the inherited text-side topology in a substantial way.

Within this setting, the results suggest that multimodal training does not collapse visual and linguistic pathways into a single undifferentiated topology; instead, it brings them into partial correspondence while preserving meaningful structural differences.

\begin{table}[t]
\centering
\caption{\textbf{Cross-modal graph alignment.}
GAUC at layer 6 of LLaVA-1.5-7B for graph embeddings matched across modalities. Higher GAUC indicates stronger structural correspondence.
}
\label{tab:matching}
\begin{tabular}{lll c}
\toprule
\textbf{$\Omega$ Modality} & \textbf{$\Gamma$ Modality} & \textbf{GAUC} \\
\midrule
Image+Text     & Image+Text     & 0.960 \\
Text           & Image          & 0.819 \\
LLaMA (Text)   & LLaVA (Text)   & 0.680 \\
\bottomrule
\end{tabular}
\end{table}

\begin{figure}[t!]
    \centering
    \includegraphics[width=\linewidth]{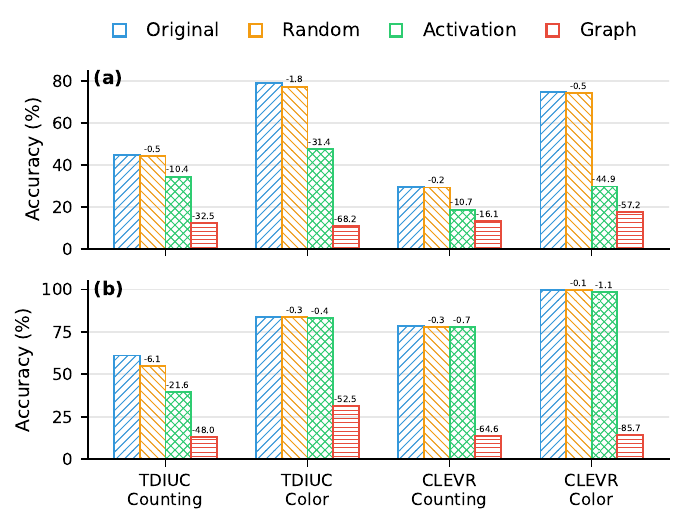}
    \caption{
        \textbf{Ablation by zeroing selected neurons.}
        Accuracy on TDIUC and CLEVR after zeroing the top 1\% of neurons selected per sample by random choice, activation magnitude, or graph degree for InternVL3-1B (a, layer 11) and Qwen2.5-VL-3B (b, layer 0). Zeroing graph-selected neurons yields the largest drop.
    }
    \label{fig:intervene_hub}
\end{figure}

\begin{figure}[t!]
    \centering
    \includegraphics[width=\linewidth]{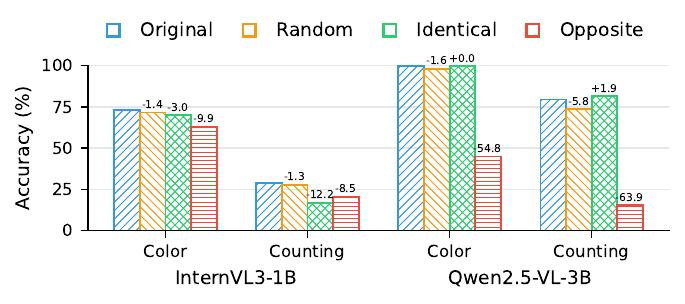}
    \caption{
        \textbf{Edge-level intervention on top-ranked neuron pairs.}
        For the strongest dataset-level graph-defined neuron pair, one endpoint is replaced by its partner’s activation (Identical), the negated partner activation (Opposite), or a random matched-shape vector (Random), and downstream accuracy is measured on color and counting tasks for InternVL3-1B and Qwen2.5-VL-3B. Opposite intervention causes the largest performance drop.
    }
    \label{fig:intervene_edge}
\end{figure}

\section{Causal Intervention Analysis}

If neural topology captures behaviorally meaningful structure, it should also identify components whose targeted perturbation materially changes model behavior. To test this, we intervene on topology-defined neurons and edges and measure the resulting performance degradation to assess whether structural centrality provides a useful criterion for selecting behaviorally influential loci.

\paragraph{Ablation of Top Neurons.} The first question is whether neurons ranked highly by graph structure are more behaviorally important than those selected by simpler criteria. For each sample, the top 1\% of neurons are ablated, chosen either by connectivity degree in the correlation graph or by activation magnitude, and the resulting performance on TDIUC and CLEVR is reported in \Cref{fig:intervene_hub}.

Across both tasks, ablating degree-ranked neurons produces the largest performance drop, indicating that graph-based ranking identifies neurons whose removal has stronger behavioral consequences than activation-based selection. The depth at which this effect is strongest differs across models: InternVL3-1B shows its largest decline around layer~11, whereas Qwen2.5-VL-3B is most sensitive at layer~0. These differences suggest that behaviorally influential topology is concentrated at different depths in different architectures, though this should not be taken as a definitive localization of multimodal fusion. Overall, structural centrality provides a more behaviorally aligned intervention criterion than activation magnitude alone.

\paragraph{Edge-Level Intervention.} Beyond individual neurons, we next test whether strong graph-defined edges encode functionally meaningful pairwise relations, rather than merely linking individually important nodes. For the edge with the highest aggregate degree across the dataset, one endpoint is intervened on while the remainder of the representation is held fixed. Specifically, its activation is replaced with that of its partner on the same edge (\textsc{Identical}), with the negated partner activation (\textsc{Opposite}), or with a random vector of matched shape (\textsc{Random}).

As shown in \Cref{fig:intervene_edge}, these interventions produce a consistent ordering of effects. The \textsc{Identical} intervention causes little degradation and can even slightly improve performance, suggesting that preserving partner-consistent activity largely maintains the underlying relation. \textsc{Random} replacement causes a moderate decline, whereas \textsc{Opposite} replacement is most destructive, especially in Qwen2.5-VL-3B, where both color and counting performance drop sharply. InternVL3-1B is more robust overall but exhibits the same qualitative ordering. This pattern indicates that the behavioral importance of a strong edge depends not only on the identity of its endpoint neurons, but also on the sign and alignment of their coordinated activity — topology is informative at the level of relations as well as nodes.

\begin{figure}[t!]
    \centering
    \includegraphics[width=\linewidth]{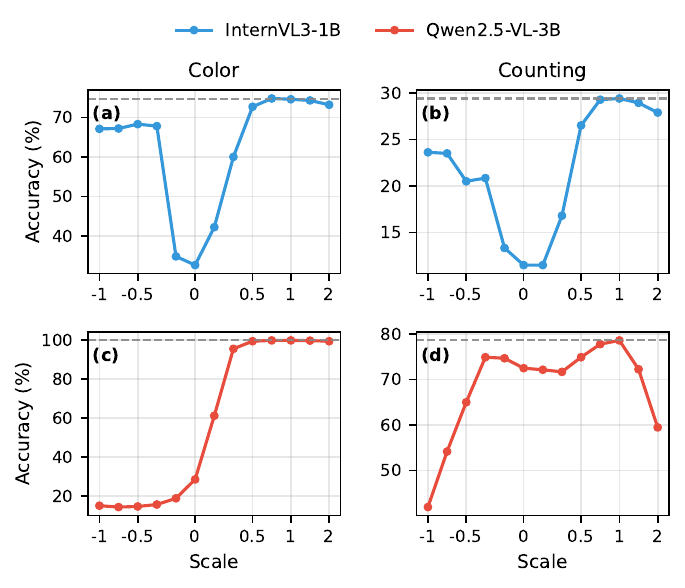}
    \caption{
        \textbf{Direct perturbation of topology-defined hub neurons.}
        Selected high-degree hub neurons in InternVL3-1B (neuron 62, layer 11) and Qwen2.5-VL-3B (neurons 71, 318, 294, 528, 583, layer 0) are scaled and evaluated on color and counting tasks. Performance degrades under both positive and negative perturbation.
    }
    \label{fig:intervene_select}
    
\end{figure}

\paragraph{Perturbation of Hub Activations.} As a final probe of causality, a small set of hub neurons is directly scaled while all other activations are held fixed: one hub in InternVL3-1B (neuron~62, layer~11) and five in Qwen2.5-VL-3B (neurons~71,~318,~294,~528,~583, layer~0), as illustrated in \Cref{fig:intervene_select}. Even small perturbations produce substantial performance degradation, and the effect is approximately symmetric under both amplification and suppression.

This symmetric sensitivity suggests that these hubs operate within a relatively narrow functional range: performance deteriorates not only when their activity is suppressed, but also when it is exaggerated. Across all intervention types, topology-defined hubs consistently occupy behaviorally influential positions. More broadly, neural topology proves useful not only for prediction and structural analysis, but also for identifying localized targets whose perturbation has substantial downstream effects.

\section{Related Work}

\paragraph{Interpretability in Vision--Language Models.}
Recent VLMs, including BLIP \cite{li2023blip}, LLaVA \cite{liu2023visual}, Qwen-VL \cite{bai2025qwen2}, and InternVL \cite{zhu2025internvl3}, achieve strong multimodal reasoning by combining visual encoders with large language models. Despite  success on instruction following and visual question answering, their internal mechanisms remain difficult to interpret \cite{dang2024explainable, lin2025survey}. Existing analyses rely on attention flow \cite{abnar2020quantifying}, saliency and gradient-based attribution \cite{selvaraju2017grad}, or interactive visualization \cite{aflalo2022vl, stan2024lvlm}. These methods provide local, token-level explanations but limited insight into the global organization of computation across layers. In parallel, mechanistic interpretability in language models has identified sparse features \cite{cunningham2023sparseautoencodershighlyinterpretable}, causal mediation pathways \cite{geiger2022inducing}, and mechanisms for editing factual associations \cite{meng2022locating}. Our work extends this structural perspective to VLMs by modeling each  layer as a neuron--neuron correlation graph and analyzing multimodal computation through its topology.

\paragraph{Neural Topology and Representation Structure.}
Prior work has studied neural networks through representational similarity \cite{kornblith2019similarity, chen2024interpretable}, neural connectivity patterns \cite{xiao2024exploring, liu2025brain, alkhamissi2025llm}, and emergent modularity in transformers \cite{zhang2023emergent}. Related mechanistic studies have also used causal interventions to test the functional importance of internal components \cite{li2023inference, rai2024practical, bartoszcze2025representation}. However, these approaches do not examine the population-level topology of VLM layers or use such structure to analyze multimodal behavior. Our framework introduces a topology-based view of VLMs by treating each layer as a neuron correlation graph and studying it through graph embeddings, modality-specific subgraphs, and causal interventions on structurally prominent neurons.

\section{Discussion}

The broader implication of this work is a shift in what should count as an explanatory unit for vision–language models. In neuroscience, understanding complex behavior required moving beyond single-neuron selectivity toward population dynamics at an intermediate, mesoscopic scale. Our results suggest an analogous perspective: within-layer co-activation topology is not merely another summary of hidden states, but a window into how computation is organized across structured populations, persistent hubs, and coordinated relations. The value of this perspective is precisely that it sits between local attribution and full circuit recovery: rich enough to expose behaviorally consequential internal organization, yet tractable enough to compare across layers, modalities, and models. Although these graphs are not literal wiring diagrams, they suggest that multimodal reasoning may be more fruitfully understood as an emergent property of organized neural populations than as the sum of independently interpretable components.

\newpage

\section*{Acknowledgments}
H.H. and Q.R.W.'s work is supported by the National Science Foundation (NSF) under Grant Nos. 2125326, 2114197, 2228533, and 2402438, as well as by the Northeastern University iSUPER Impact Engine.
Any opinions, findings, conclusions, or recommendations expressed in the paper are those of the authors and do not necessarily reflect the views of the funding agencies.

{
    \small
    \bibliographystyle{ieeenat_fullname}
    \bibliography{main}
}

\clearpage
\setcounter{page}{1}
\maketitlesupplementary

\section{Additional Experimental Setup}
\label{app:setup}

\paragraph{Models.} We evaluate three representative vision--language models: InternVL3-1B~\cite{zhu2025internvl3}, Qwen2.5-VL Instruct-3B~\cite{bai2025qwen2}, and LLaVA-1.5-7B~\cite{liu2023visual}. These models span both instruction-tuned and general-purpose VLM families. InternVL3 uses an EVA-style visual encoder~\cite{sun2023evaclipimprovedtrainingtechniques}, Qwen2.5-VL adopts a dynamic-resolution visual frontend, and LLaVA-1.5 uses CLIP ViT-L/14 as its visual encoder. For each model, we extract layerwise hidden activations from the multimodal transformer backbone after projected visual tokens are concatenated with tokenized text embeddings, and construct correlation graphs from these hidden states.

\paragraph{Datasets.} We evaluate predictability on seven benchmarks: CLEVR~\cite{johnson2017clevr}, TDIUC~\cite{kafle2017analysis}, MMMU~\cite{yue2024mmmu}, MMMU-Pro~\cite{yue2025mmmu}, BLINK~\cite{fu2024blink}, EMMA~\cite{hao2025can}, and MHaluBench~\cite{chen2024unified}.

For CLEVR, we adapt the task to object counting using 10{,}000 image--question pairs, where the model is asked to count the number of visible objects in a synthetic scene. For TDIUC, we use the sports recognition subset containing 4{,}634 image--question pairs across six activity categories: baseball, surfing, skiing, tennis, frisbee, and skateboarding. For MHaluBench, we use the validation split with 2{,}110 examples, consisting of 1{,}055 hallucinating and 1{,}055 non-hallucinating responses. Each sample is formed by concatenating a question with either a faithful or hallucinated answer. CLEVR, TDIUC, and MHaluBench are emphasized in the main paper because they provide focused tests of numerical grounding, semantic discrimination, and multimodal factual consistency, respectively. Results on MMMU, MMMU-Pro, BLINK, and EMMA are included to assess broader generality.

\paragraph{Training Details.} For each dataset, we randomly split examples into 80\% training and 20\% test sets. We train both a linear probe and a GCN probe on each layer's graph representation using cross-entropy loss for classification and standard regression objectives for numerical prediction. Optimization is performed with Adam~\cite{kingma2014adam}, and we report the best test performance across epochs. All experiments are conducted on a single NVIDIA L40S GPU.

\paragraph{Text-only Baselines for Hallucination.} For hallucination detection on MHaluBench, we construct two text-only baselines using \texttt{word2vec}~\cite{mikolov2013efficientestimationwordrepresentations, rehurek2010gensim}: the mean embedding of the question--answer prompt and the token count of the prompt. Separate linear classifiers are trained on these features to quantify how much hallucination is predictable from shallow textual information alone before introducing graph-based multimodal representations.

\begin{table*}[b]
\centering
\small
\caption{
\textbf{Precision and recall for graph probing across multimodal benchmarks.}
Companion to \Cref{tab:vlm_cls}, reporting Precision and Recall for linear and graph-based probes on TDIUC, CLEVR, MMMU, MMMU-Pro, BLINK, and EMMA. Best result in each model–dataset pair is in \textbf{bold}.
}
\label{tab:vlm_cls_pr}
\begin{tabular}{lcccccccccccc}
\toprule
& \multicolumn{4}{c}{\textbf{InternVL3-1B}} 
& \multicolumn{4}{c}{\textbf{Qwen2.5-VL-3B}} 
& \multicolumn{4}{c}{\textbf{LLaVA-1.5-7B}} \\
\cmidrule(lr){2-5} \cmidrule(lr){6-9} \cmidrule(lr){10-13}
\textbf{Dataset}
& \multicolumn{2}{c}{\textbf{Linear}} & \multicolumn{2}{c}{\textbf{GCN}}
& \multicolumn{2}{c}{\textbf{Linear}} & \multicolumn{2}{c}{\textbf{GCN}}
& \multicolumn{2}{c}{\textbf{Linear}} & \multicolumn{2}{c}{\textbf{GCN}} \\
\cmidrule(lr){2-3} \cmidrule(lr){4-5}
\cmidrule(lr){6-7} \cmidrule(lr){8-9}
\cmidrule(lr){10-11} \cmidrule(lr){12-13}
& Prec & Rec & Prec & Rec & Prec & Rec & Prec & Rec & Prec & Rec & Prec & Rec \\
\midrule
TDIUC
& 0.879 & 0.839 & \textbf{0.961} & \textbf{0.957}
& 0.932 & 0.929 & \textbf{0.970} & \textbf{0.970}
& \textbf{0.964} & \textbf{0.967} & 0.945 & 0.949 \\
CLEVR
& 0.980 & 0.980 & \textbf{0.993} & \textbf{0.993}
& 0.919 & 0.919 & \textbf{0.962} & \textbf{0.963}
& 0.619 & 0.594 & \textbf{0.712} & \textbf{0.684} \\
MMMU
& \textbf{0.270} & 0.288 & 0.231 & \textbf{0.300}
& 0.226 & 0.279 & \textbf{0.337} & \textbf{0.337}
& 0.320 & \textbf{0.324} & \textbf{0.389} & 0.261 \\
MMMU-Pro
& \textbf{0.368} & 0.280 & 0.272 & \textbf{0.323}
& 0.157 & 0.256 & \textbf{0.299} & \textbf{0.300}
& 0.224 & 0.260 & \textbf{0.306} & \textbf{0.306} \\
BLINK
& 0.547 & 0.541 & \textbf{0.591} & \textbf{0.590}
& 0.544 & 0.544 & \textbf{0.564} & \textbf{0.564}
& \textbf{0.653} & \textbf{0.650} & 0.592 & 0.591 \\
EMMA
& 0.160 & 0.288 & \textbf{0.332} & \textbf{0.322}
& 0.167 & 0.247 & \textbf{0.316} & \textbf{0.324}
& 0.359 & 0.267 & \textbf{0.386} & \textbf{0.296} \\
\bottomrule
\end{tabular}
\end{table*}

\end{document}